\documentclass[11pt]{article}

\usepackage[utf8]{inputenc}
\usepackage[T1]{fontenc}
\usepackage{amsmath,amsfonts}
\usepackage{graphicx}
\usepackage{booktabs}
\usepackage{microtype}
\usepackage{nicefrac}
\usepackage{xcolor}
\usepackage{url}
\usepackage[numbers]{natbib}    
\usepackage{hyperref}

\title{Physics-Guided Machine Learning for Uncertainty Quantification in Turbulence Models}

\author{
  Minghan Chu\thanks{Also with the Institute of Mechanics, Chinese Academy of Sciences.} \\
  Department of Mechanical and Materials Engineering, Queen’s University \\
  \texttt{17mc93@queensu.ca} \\
  \texttt{chuminghan@imech.ac.cn}
  \and
  Weicheng Qian \\
  Department of Computer Science, University of Saskatchewan \\
  \texttt{weicheng.qian@usask.ca}
}

\date{} 

\begin{document}
\maketitle

\begin{center}
\textbf{This work was accepted at the NeurIPS 2025 Workshop on Machine Learning for the Physical Sciences (ML4PS).}
\end{center}
\vspace{0.5em}

\begin{abstract}
Predicting the evolution of turbulent flows is central across science and engineering. Most studies rely on simulations with turbulence models, whose empirical simplifications introduce epistemic uncertainty. The Eigenspace Perturbation Method (EPM) is a widely used physics-based approach to quantify model-form uncertainty, but being purely physics-based it can overpredict uncertainty bounds. We propose a convolutional neural network (CNN)–based modulation of EPM perturbation magnitudes to improve calibration while preserving physical consistency. Across canonical cases, the hybrid ML–EPM framework yields substantially tighter, better-calibrated uncertainty estimates than baseline EPM alone.
\end{abstract}

\section{Introduction}

Turbulent flows govern the transport of momentum, heat, and mass in natural and engineering systems—from planetary atmospheres to aerodynamic surfaces. Despite their ubiquity, turbulence remains difficult to predict accurately because of the need to model unresolved scales, leading to the long-standing closure problem. Engineering simulations rely on turbulence models to approximate these effects; however, their empirical nature introduces significant epistemic uncertainty in predictions. Understanding and reducing this uncertainty remains a grand scientific challenge, central to advances in energy efficiency, environmental forecasting, and the optimization of industrial and aerospace designs.

As we do not have an analytical theory to predict turbulent flow evolution, engineering and scientific studies have to rely on computer simulations \citep{moin1997tackling}. Exact simulation of the evolution of turbulent flows, known as Direct Numerical Simulation (DNS) \citep{moin1998direct, moser1999direct, lee2015direct}, is computationally intractable for almost all engineering and scientific problems. DNS requires resolving the full spectrum of turbulent motion. For the high Reynolds number flows common in real-life problems, the range of these scales is so large that the required computational grid would exceed the capacity of even the most powerful supercomputers, for now and in the foreseeable future. Turbulence modeling provides a necessary compromise: instead of resolving every eddy, turbulence models try to represent the overall effect of the smaller-scale turbulent fluctuations on the larger flow field. This leads to the critical closure problem. When the governing Navier–Stokes equations are averaged to make them computationally tractable using Reynolds averaging, new, unknown terms representing the turbulent stresses emerge, referred to as Reynolds stresses. A turbulence model is an approximate constitutive relation providing a closure for these unknown terms by relating them to the mean flow. There are different fidelities of turbulence modeling approaches, ranging from sub-grid scale models for Large Eddy Simulations (LES) \citep{lesieur1996new, lesieur2005large}, Reynolds Stress Modeling (RSM) \citep{speziale1991analytical, launder1975progress, mishra2017toward,speziale1991modelling}, Eddy Viscosity Modeling (EVM) \cite{craft1996development}, etc. The ability to predict aerodynamic forces, heat transfer rates, mixing efficiencies, etc.\ hinges on the quality of the model used to approximate turbulence physics. Due to their lower computational cost, most engineering applications utilize simple EVMs such as the $k$–$\epsilon$ \citep{launder1983numerical} and $k$–$\omega$ \citep{wilcox2008formulation} models, where the Boussinesq and Gradient Diffusion Hypotheses are used to render the final model tractable \citep{duraisamy2019turbulence,  mishra2016sensitivity, oliver2009uncertainty}. 

Uncertainty quantification (UQ) \citep{smith2013uncertainty} is essential for turbulence modeling because every turbulence model is an incomplete and empirical approximation of true turbulence physics. Turbulence models introduce closure terms to represent the effects of unresolved turbulent scales, but the mathematical forms of these closures are inexact. This introduces epistemic uncertainty in model predictions because the turbulence model’s underlying assumptions are known to be invalid for complex real-life turbulent flows, such as those with significant streamline curvature, adverse pressure gradients, or flow separation. A deterministic computational fluid dynamics (CFD) simulation using a turbulence model provides a single prediction that may have large errors and uncertainties. Thus, UQ provides the necessary framework to move beyond a single, deterministic prediction to a more informative probabilistic prediction. For engineering design and analysis, it allows for the rational estimation of design margins of safety and risk assessment \citep{wood1990modeling, du2000methodology, gurnani2005robust}. For scientific studies, turbulence-model UQ helps to distinguish between genuine physical phenomena predicted by a simulation and artifacts that may arise due to turbulence-model errors.

The Eigenspace Perturbation Framework (EPM) \citep{iaccarino2017eigenspace} is a widely used method to predict turbulence-model uncertainties. This purely physics-based method relies on perturbations introduced to the shape, alignment and size (i.e., the eigenvalues, eigenvectors and amplitude) of the modeled Reynolds-stress ellipsoid. The magnitudes of these perturbations are determined from physics principles. The EPM has been applied across engineering design and analysis in the recent past with substantial success and is a de facto standard for UQ in turbulence \citep{mishra2019uncertainty, mishra2017rans, mishra2019estimating, mishra2017uncertainty, thompson2019eigenvector, demir2023robust, cook2019optimization, mishra2020design, righi2023uncertainties, li2024adjoint, li2025application, huang2020nonuniform, mukhopadhaya2020multi}. But it suffers from shortcomings due to its reliance on physics principles only. The perturbations are dependent on the uncertainty in the turbulence model’s predictions. This uncertainty varies across different turbulent flows and also across different regions of the same turbulent flow. Based on physics principles alone, the EPM can only determine the maximal physically permissible uncertainty. This often leads to over-generous and imprecisely calibrated uncertainty bounds. This can have a cascading effect on design and analysis, for instance leading to over-conservative final designs with a high factor of safety. To improve the EPM, the turbulence modeling community needs a functionality to modulate the magnitude of these perturbations so as to make them reflect the degree of prediction error by the turbulence model in the specific turbulent flow being simulated and at the particular location in the flow domain. Such a function may not be accessible using only physics principles. However, we can utilize neural networks which, as universal approximators, may be able to model this mapping from data. In this investigation, we use convolutional neural networks (CNNs) to act as surrogate models for this function. In our methodology, using paired datasets of high-fidelity simulations and turbulence-model predictions, the model learns to predict the discrepancy between turbulence-model predictions and the ground truth. This learned mapping serves as a predictor for the magnitude of perturbations in the EPM. We use this learned mapping to correct the turbulence-model prediction. Using the trained model to modulate the EPM’s perturbation magnitudes represents a physics-guided machine learning approach \citep{duraisamy2019turbulence, brunton2020machine, chung2021data, chung2022interpretable, duraisamy2021perspectives, hoidn2023physics} where physical principles answer \emph{how to perturb} and the machine-learning model governs \emph{how much to perturb} to improve turbulence modeling. 

\section{Methodology and theoretical background}

This section provides a concise yet complete description of the proposed hybrid ML–EPM framework, outlining both the theoretical basis and the data-driven implementation. Turbulent flows exhibit chaotic and multiscale behavior in both space and time. The instantaneous velocity field $\tilde{u}_i$ can be decomposed into a mean and a fluctuating component using Reynolds averaging \citep{pope2001turbulent}:
\begin{equation}
\tilde{u}_i = U_i + u_i,
\end{equation}
where $U_i$ is the mean velocity and $u_i$ the fluctuating velocity with zero mean. The covariance of these fluctuations defines the Reynolds-stress tensor, $R_{ij} = \left\langle u_i u_j \right\rangle$, which embodies the closure problem of turbulence modeling: predicting $R_{ij}$ requires modeling unknown higher-order correlations.  

Being symmetric and positive semi-definite, $R_{ij}$ can be decomposed into its eigenspace representation as
\begin{equation}
R_{ij} = 2\rho k \left( v_{in}\Lambda_{nl}v_{lj} + \tfrac{1}{3}\delta_{ij} \right),
\end{equation}
where $k$ is the turbulent kinetic energy (TKE), $\Lambda$ the diagonal matrix of eigenvalues, and $v$ the eigenvector matrix. The Eigenspace Perturbation Method (EPM) introduces physically constrained perturbations in $k$, $\Lambda$, and $v$ to explore model-form uncertainty while ensuring realizability. The perturbed tensor reads
\begin{equation}
R_{ij}^* = 2\rho k^* \left( v_{in}^*\Lambda_{nl}^*v_{lj}^* + \tfrac{1}{3}\delta_{ij} \right),
\end{equation}
where the asterisk denotes perturbed quantities. In this study, the data-driven correction is applied to $k$, while the RANS-predicted anisotropy structure ($v_{ij}$, $\Lambda_{ij}$) is retained to preserve physical consistency.

\begin{figure}[t]
    \centering
    \includegraphics[width=0.7\linewidth]{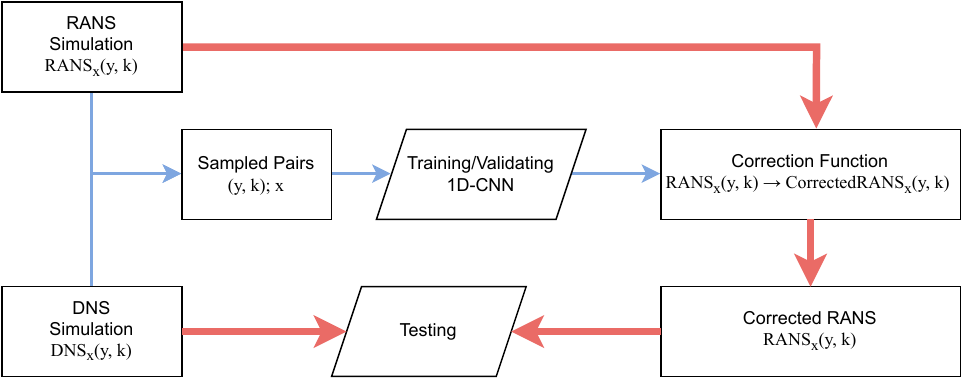}
    \caption{Data flow and methodology of the proposed framework. The blue path denotes the training stage using paired RANS–DNS data, and the red path indicates the validation and testing stage.}
    \label{fig:data-flow}
\end{figure}

The turbulent kinetic energy fields from RANS and DNS are expressed as spatial mappings
\begin{equation}
k^{*} = f(x,y),
\end{equation}
where $(x,y)$ are coordinates in the two-dimensional computational domain. The correction function $Z$ transforms the RANS-predicted TKE field to its DNS counterpart:
\begin{equation}
Z: f^{\mathrm{RANS}}(x,y) \rightarrow f^{\mathrm{DNS}}(x,y),
\end{equation}
or equivalently as a pointwise mapping
\begin{equation}
\zeta: (x, y, k^{\mathrm{RANS}}) \rightarrow (x, y, k^{\mathrm{DNS}}).
\end{equation}
This formulation treats the TKE correction as a supervised learning task: given paired RANS and DNS samples $\big(\mathbf{k}^{\mathrm{RANS}}_{x,y,\delta}, \mathbf{k}^{\mathrm{DNS}}_{x,y,\delta}\big)$, the model learns a nonlinear function $\hat{g}$ such that
\begin{equation}
\hat{k}^{\mathrm{DNS}} = \hat{g}\!\left(k^{\mathrm{RANS}}; \theta\right),
\end{equation}
where $\theta$ are trainable parameters obtained by minimizing the mean-squared error
\begin{equation}
\mathcal{L}(\theta) = \frac{1}{N}\sum_{i=1}^{N} \left[\hat{g}\!\left(k_i^{\mathrm{RANS}}\right) - k_i^{\mathrm{DNS}}\right]^2.
\end{equation}

\subsection{Neural network architecture and training}

A lightweight one-dimensional convolutional neural network (1D-CNN) is employed to learn the mapping $\hat{g}$. The network comprises two convolutional layers (kernel size 3) followed by a max-pooling operation and two fully connected layers, totaling approximately $86$ parameters. Each convolutional layer uses ReLU activation and batch normalization to stabilize training. The compact design is motivated by the limited size of the training data and the desire for interpretability.  

The model is trained using the Adam optimizer with a learning rate of $10^{-3}$ and the Mean Absolute Error (MAE) loss function. Training, validation, and test splits are set to $75\%$, $5\%$, and $20\%$, respectively. Early stopping with a patience of 10 epochs prevents overfitting. The training and validation data come from two canonical turbulent flow configurations: (i) the SD7003 airfoil \citep{catalano2010turbulence} and (ii) the periodic hill \citep{rapp2011flow}. These cases include adverse pressure gradients, streamline curvature, and flow separation—conditions that often challenge RANS turbulence models.

\subsection{Integration into the EPM Framework}

The learned correction $\hat{g}$ is integrated into the EPM framework by replacing the RANS-predicted TKE with its corrected counterpart in the Reynolds-stress reconstruction:
\begin{equation}
R_{ij}^{\mathrm{corr}} = 2\,\hat{k}^{\mathrm{DNS}}\, b_{ij}^{\mathrm{RANS}},
\end{equation}
where $b_{ij}^{\mathrm{RANS}}$ is the normalized anisotropy tensor. This approach modifies only the magnitude of turbulence energy while preserving anisotropy directions and realizability. Thus, the proposed ML–EPM coupling remains consistent with turbulence physics while enabling data-driven refinement of model predictions.

\begin{figure}[t]
    \centering
    \includegraphics[width=\linewidth]{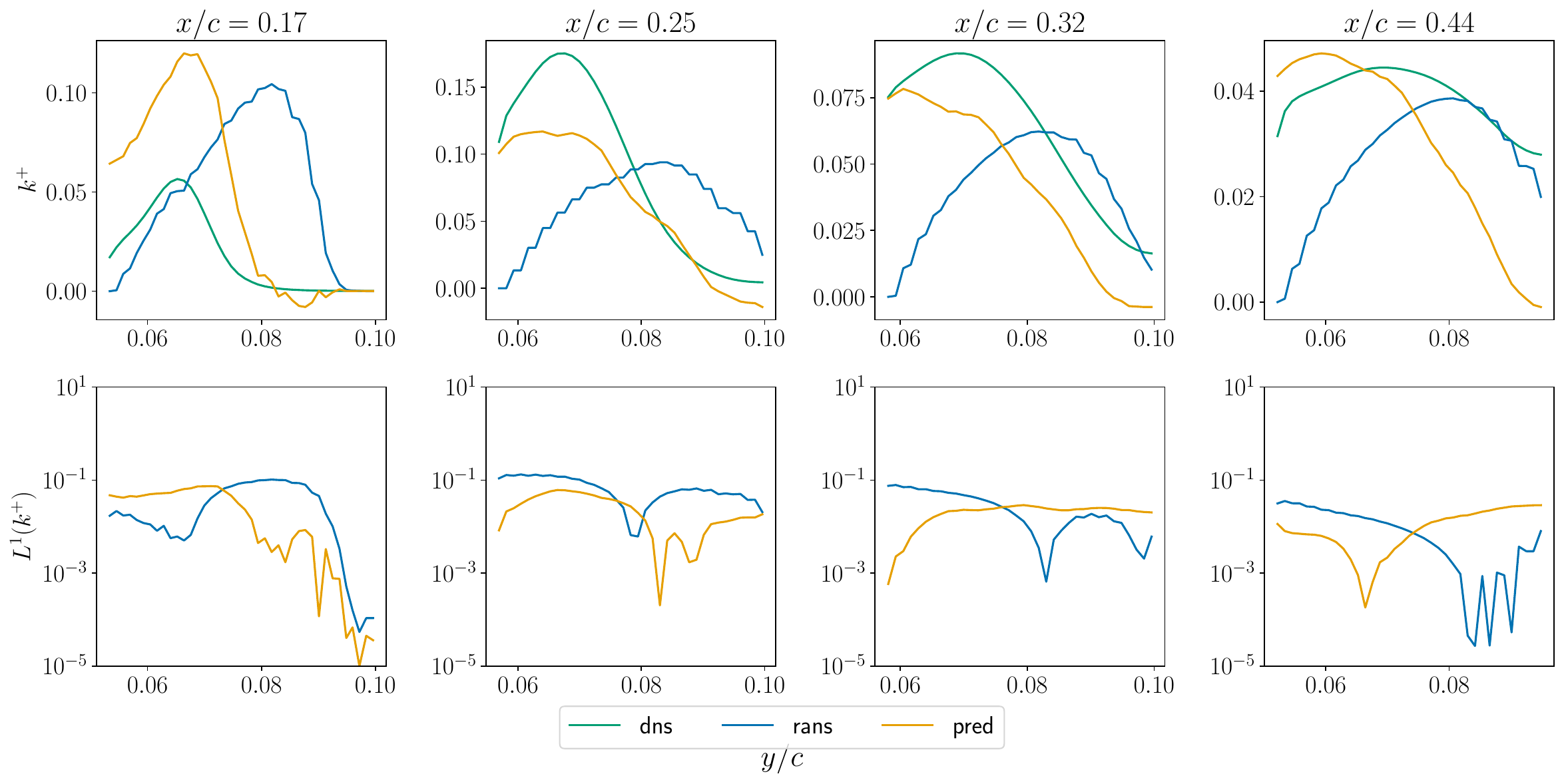}
    \caption{Illustration for the SD7003 airfoil case. The first row compares CNN predictions, baseline RANS, and DNS data. The second row shows the MAE of CNN-corrected and baseline predictions at different chordwise positions.}
    \label{fig:sd7003}
\end{figure}

\section{Results}

In Figure~\ref{fig:sd7003}, we show results of the trained CNN model for turbulent flow over an SD7003 airfoil, focusing on normalized turbulent kinetic energy profiles $k^{+}$ (normalized with respect to the freestream velocity) at different locations along the airfoil. The first row reports the predictions of the CNN, the turbulence model and ground-truth DNS data. The second row reports the corresponding MAE for the turbulence model and the CNN, with DNS as ground truth. As can be seen in the first row, the CNN-predicted profiles (orange) are in better agreement with DNS profiles (green) than the turbulence-model predictions (blue). Examining the MAE in the second row, it is clear that the CNN reduces the error between the turbulence-model predictions and DNS by one to two orders of magnitude. This represents a significant increase in prediction accuracy. 

\begin{figure}  
    \centering
    \includegraphics[width=\linewidth]{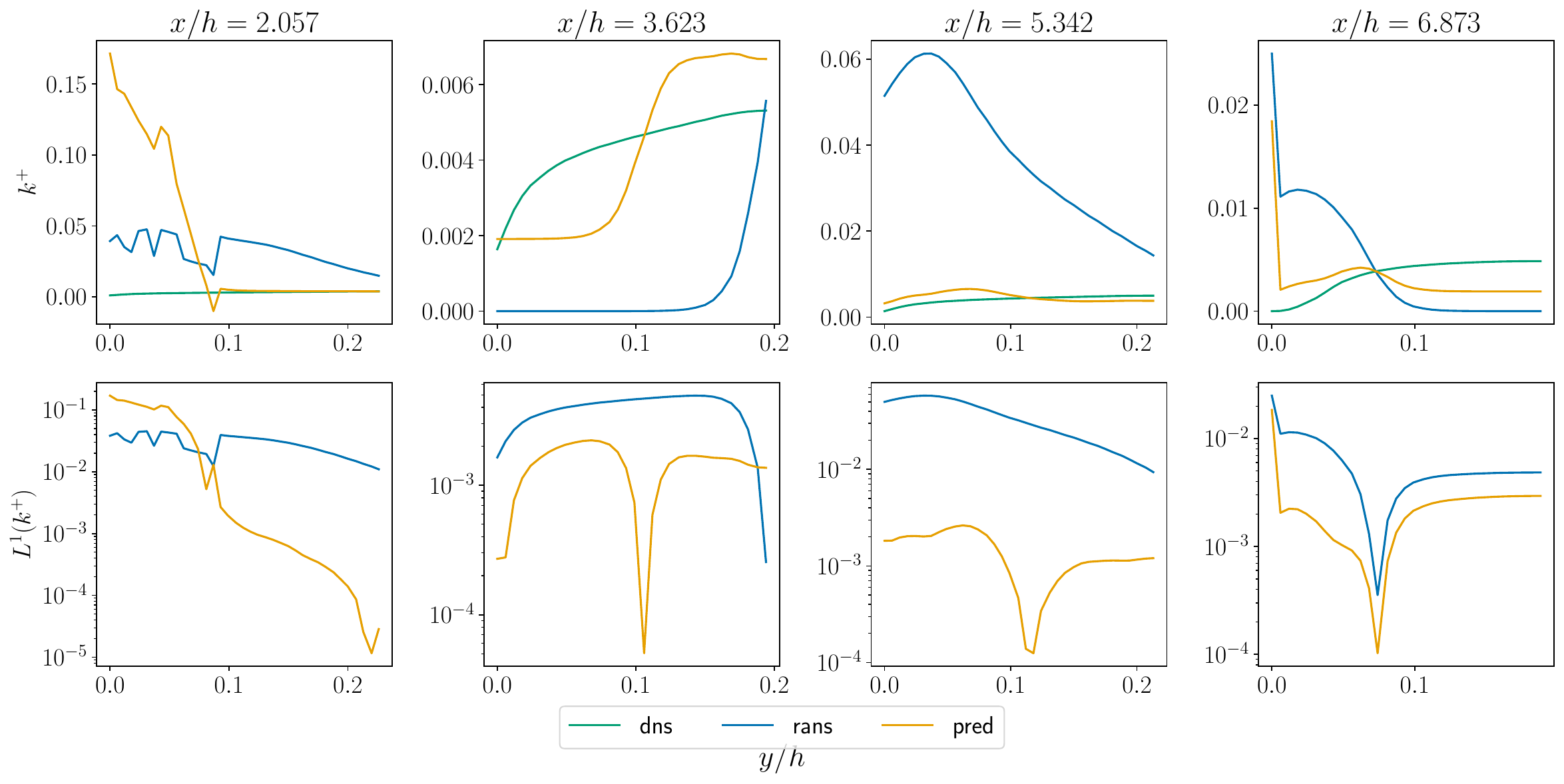}
    \caption{Results for turbulent flow over periodic hills: Top—CNN, RANS, and DNS; Bottom—MAE for CNN vs.\ RANS at multiple $x/h$ stations (streamwise position normalized by hill height).}
    \label{fig:phill}
\end{figure}

In Figure~\ref{fig:phill}, we report corresponding results for a single case from the paired dataset corresponding to turbulent flow over periodic hills. The results are for a flow case not used in training, so there is no risk of data bleed. Here, we have turbulent flow separation with a re-attachment at $x/h = 4.769$. We can see that at $x/h = 2.057$, which lies inside the separation bubble, the CNN-corrected turbulent kinetic profile is massively improved from the baseline turbulence-model prediction. This is also the case around the re-attachment point at $x/h = 5.342$. The CNN-corrected turbulent kinetic energy profile is almost coincident with the DNS result, while the turbulence-model prediction is patently incorrect. The error on the CNN-corrected profile is correspondingly almost three orders of magnitude lower. Further downstream in the fully developed boundary layer, the turbulence model has lower errors that are still higher than the CNN-corrected error. In this case as well, the CNN-corrected solutions are one to two orders of magnitude more accurate than the baseline model. Compared with purely data-driven correction approaches (e.g., field-inversion neural networks \citep{parish2016paradigm}), the present ML–EPM framework achieves comparable error reduction while retaining full physical interpretability through its eigenspace formulation.

\paragraph{Validation, limitations, and discussion.}
The proposed ML–EPM framework is evaluated on two canonical turbulent flows—the SD7003 airfoil and the periodic hill—which capture key physical phenomena such as curvature, separation, and reattachment. Despite the limited dataset, these cases are representative of conditions that challenge traditional turbulence models. The improved performance arises from the physics-guided nature of the ML–EPM coupling: the CNN modulates only the turbulence kinetic-energy magnitude while preserving the anisotropy structure predicted by RANS, thereby preventing unphysical stress distortions common in purely data-driven methods. Quantitatively, the mean absolute error (MAE) between predicted and DNS fields is reduced by one to two orders of magnitude compared with baseline RANS predictions across all test locations. Nevertheless, the restricted dataset limits statistical diversity, and the model’s generalization to three-dimensional or higher-Reynolds-number configurations remains to be tested. Future work will extend validation to additional geometries and flow regimes to further assess robustness and scalability.

\section{Conclusions}
The objective of this investigation is to develop a hybrid physics-guided machine-learning methodology for uncertainty and error estimation in turbulence-model predictions. Prior studies have explored ML within the EPM framework \cite{heyse2021data, heyse2021estimating, matha2023evaluation}. We augment EPM with a trained CNN: the physics-based EPM guides \emph{how} perturbations are injected, while the ML model guides \emph{how much} to perturb, yielding calibrated uncertainty estimates. On SD7003 and periodic-hill benchmarks using paired DNS and turbulence-model data, the CNN reduces error by one to two orders of magnitude relative to baseline RANS.

\bibliographystyle{plain}
\bibliography{bibliography}

@PREAMBLE{
 "\providecommand{\noopsort}[1]{}" 
 # "\providecommand{\singleletter}[1]{#1}%" 
}

@article{hoidn2023physics,
  title={Physics constrained unsupervised deep learning for rapid, high resolution scanning coherent diffraction reconstruction},
  author={Hoidn, Oliver and Mishra, Aashwin Ananda and Mehta, Apurva},
  journal={Scientific reports},
  volume={13},
  number={1},
  pages={22789},
  year={2023},
  publisher={Nature Publishing Group UK London}
}

@incollection{launder1983numerical,
  title={The numerical computation of turbulent flows},
  author={Launder, Brian Edward and Spalding, Dudley Brian},
  booktitle={Numerical prediction of flow, heat transfer, turbulence and combustion},
  pages={96--116},
  year={1983},
  publisher={Elsevier}
}

@article{mishra2017toward,
  title={Toward approximating non-local dynamics in single-point pressure--strain correlation closures},
  author={Mishra, Aashwin A and Girimaji, Sharath S},
  journal={Journal of Fluid Mechanics},
  volume={811},
  pages={168--188},
  year={2017},
  publisher={Cambridge University Press}
}

@article{wilcox2008formulation,
  title={Formulation of the kw turbulence model revisited},
  author={Wilcox, David C},
  journal={AIAA journal},
  volume={46},
  number={11},
  pages={2823--2838},
  year={2008}
}

@book{smith2013uncertainty,
  title={Uncertainty quantification: theory, implementation, and applications},
  author={Smith, Ralph C},
  volume={12},
  year={2013},
  publisher={Siam}
}

@article{matha2023evaluation,
  title={Evaluation of physics constrained data-driven methods for turbulence model uncertainty quantification},
  author={Matha, Marcel and Kucharczyk, Karsten and Morsbach, Christian},
  journal={Computers \& Fluids},
  volume={255},
  pages={105837},
  year={2023},
  publisher={Elsevier}
}

@article{demir2023robust,
  title={Robust shape optimization under model uncertainty of an aircraft wing using proper orthogonal decomposition and inductive design exploration method},
  author={Demir, Gorkem and Gorguluarslan, Recep M and Aradag, Selin},
  journal={Structural and Multidisciplinary Optimization},
  volume={66},
  number={4},
  pages={93},
  year={2023},
  publisher={Springer}
}

@article{mukhopadhaya2020multi,
  title={Multi-fidelity modeling of probabilistic aerodynamic databases for use in aerospace engineering},
  author={Mukhopadhaya, Jayant and Whitehead, Brian T and Quindlen, John F and Alonso, Juan J and Cary, Andrew W},
  journal={International Journal for Uncertainty Quantification},
  volume={10},
  number={5},
  year={2020},
  publisher={Begel House Inc.}
}

@inproceedings{righi2023uncertainties,
  title={Uncertainties Quantification in the Prediction of the Aeroelastic Response of The PAZY Wing Tunnel Model},
  author={Righi, Marcello},
  booktitle={AIAA SCITECH 2023 Forum},
  pages={0761},
  year={2023}
}

@article{huang2020nonuniform,
  title={A nonuniform perturbation to quantify RANS model uncertainties},
  author={Huang, Z and Mishra, A and Iaccarino, G},
  journal={Center for Turbulence Research Annual Research Briefs, Stanford Univ., Stanford, CA},
  pages={223--232},
  year={2020}
}

@article{li2025application,
  title={Application of Uncertainty Quantification of Reynolds-Averaged Navier--Stokes Models in Hypersonic Flow},
  author={Li, Anna and Wang, Tongsheng and Chen, Jianan and Xi, Guang and Huang, Zhu},
  journal={AIAA Journal},
  pages={1--9},
  year={2025},
  publisher={American Institute of Aeronautics and Astronautics}
}

@article{li2024adjoint,
  title={Adjoint design optimization under the uncertainty quantification of Reynolds-Averaged Navier-Stokes turbulence model},
  author={Li, Anna and Wang, Tongsheng and Chen, Jianan and Huang, Zhu and Xi, Guang},
  journal={AIAA Journal},
  volume={62},
  number={7},
  pages={2589--2600},
  year={2024},
  publisher={American Institute of Aeronautics and Astronautics}
}

@misc{pope2001turbulent,
  title={Turbulent flows},
  author={Pope, Stephen B},
  year={2001},
  publisher={IOP Publishing}
}

@article{iaccarino2017eigenspace,
  title={Eigenspace perturbations for uncertainty estimation of single-point turbulence closures},
  author={Iaccarino, Gianluca and Mishra, Aashwin Ananda and Ghili, Saman},
  journal={Physical Review Fluids},
  volume={2},
  number={2},
  pages={024605},
  year={2017},
  publisher={APS}
}

@article{mishra2017rans,
  title={RANS predictions for high-speed flows using enveloping models},
  author={Mishra, AA and Iaccarino, G},
  journal={arXiv preprint arXiv:1704.01699},
  year={2017}
}

@article{mishra2019uncertainty,
  title={Uncertainty estimation module for turbulence model predictions in SU2},
  author={Mishra, Aashwin Ananda and Mukhopadhaya, Jayant and Iaccarino, Gianluca and Alonso, Juan},
  journal={AIAA Journal},
  volume={57},
  number={3},
  pages={1066--1077},
  year={2019},
  publisher={American Institute of Aeronautics and Astronautics}
}

@article{mishra2016sensitivity,
  title={Sensitivity of flow evolution on turbulence structure},
  author={Mishra, Aashwin A and Iaccarino, Gianluca and Duraisamy, Karthik},
  journal={Physical Review Fluids},
  volume={1},
  number={5},
  pages={052402},
  year={2016},
  publisher={APS}
}

@article{craft1996development,
  title={Development and application of a cubic eddy-viscosity model of turbulence},
  author={Craft, TJ and Launder, BE and Suga, K},
  journal={International Journal of Heat and Fluid Flow},
  volume={17},
  number={2},
  pages={108--115},
  year={1996},
  publisher={Elsevier}
}

@article{duraisamy2019turbulence,
  title={Turbulence modeling in the age of data},
  author={Duraisamy, Karthik and Iaccarino, Gianluca and Xiao, Heng},
  journal={Annual Review of Fluid Mechanics},
  volume={51},
  pages={357--377},
  year={2019},
  publisher={Annual Reviews}
}

@article{parish2016paradigm,
  title={A paradigm for data-driven predictive modeling using field inversion and machine learning},
  author={Parish, Eric J and Duraisamy, Karthik},
  journal={Journal of Computational Physics},
  volume={305},
  pages={758--774},
  year={2016},
  publisher={Elsevier}
}

@inproceedings{oliver2009uncertainty,
  title={Uncertainty quantification for RANS turbulence model predictions},
  author={Oliver, Todd and Moser, Robert},
  booktitle={APS division of fluid dynamics meeting abstracts},
  volume={62},
  pages={LC--004},
  year={2009}
}

@article{thompson2019eigenvector,
  title={Eigenvector perturbation methodology for uncertainty quantification of turbulence models},
  author={Thompson, Roney L and Mishra, Aashwin Ananda and Iaccarino, Gianluca and Edeling, Wouter and Sampaio, Luiz},
  journal={Physical Review Fluids},
  volume={4},
  number={4},
  pages={044603},
  year={2019},
  publisher={APS}
}

@article{cook2019optimization,
  title={Optimization under turbulence model uncertainty for aerospace design},
  author={Cook, Laurence W and Mishra, AA and Jarrett, JP and Willcox, KE and Iaccarino, G},
  journal={Physics of Fluids},
  volume={31},
  number={10},
  pages={105111},
  year={2019},
  publisher={AIP Publishing LLC}
}

@article{mishra2020design,
  title={Design exploration and optimization under uncertainty},
  author={Mishra, Aashwin Ananda and Mukhopadhaya, Jayant and Alonso, Juan and Iaccarino, Gianluca},
  journal={Physics of Fluids},
  volume={32},
  number={8},
  pages={085106},
  year={2020},
  publisher={AIP Publishing LLC}
}

@article{heyse2021estimating,
  title={Estimating RANS model uncertainty using machine learning},
  author={Heyse, Jan Felix and Mishra, Aashwin A and Iaccarino, Gianluca},
  journal={Journal of the Global Power and Propulsion Society},
  volume={2021},
  number={May},
  pages={1--14},
  year={2021},
  publisher={Global Power and Propulsion Society}
}

@inproceedings{heyse2021data,
  title={Data Driven Physics Constrained Perturbations for Turbulence Model Uncertainty Estimation.},
  author={Heyse, Jan Felix and Mishra, Aashwin Ananda and Iaccarino, Gianluca},
  booktitle={AAAI Spring Symposium: MLPS},
  year={2021}
}

@article{mishra2017uncertainty,
  title={Uncertainty estimation for reynolds-averaged navier--stokes predictions of high-speed aircraft nozzle jets},
  author={Mishra, Aashwin Ananda and Iaccarino, Gianluca},
  journal={AIAA Journal},
  volume={55},
  number={11},
  pages={3999--4004},
  year={2017},
  publisher={American Institute of Aeronautics and Astronautics}
}

@article{mishra2019estimating,
  title={Estimating uncertainty in homogeneous turbulence evolution due to coarse-graining},
  author={Mishra, Aashwin Ananda and Duraisamy, Karthik and Iaccarino, Gianluca},
  journal={Physics of Fluids},
  volume={31},
  number={2},
  pages={025106},
  year={2019},
  publisher={AIP Publishing LLC}
}

@article{chung2022interpretable,
  title={Interpretable data-driven methods for subgrid-scale closure in LES for transcritical LOX/GCH4 combustion},
  author={Chung, Wai Tong and Mishra, Aashwin Ananda and Ihme, Matthias},
  journal={Combustion and Flame},
  volume={239},
  pages={111758},
  year={2022},
  publisher={Elsevier}
}

@article{chung2021data,
  title={Data-assisted combustion simulations with dynamic submodel assignment using random forests},
  author={Chung, Wai Tong and Mishra, Aashwin Ananda and Perakis, Nikolaos and Ihme, Matthias},
  journal={Combustion and Flame},
  volume={227},
  pages={172--185},
  year={2021},
  publisher={Elsevier}
}

@article{brunton2020machine,
  title={Machine learning for fluid mechanics},
  author={Brunton, Steven L and Noack, Bernd R and Koumoutsakos, Petros},
  journal={Annual Review of Fluid Mechanics},
  volume={52},
  pages={477--508},
  year={2020},
  publisher={Annual Reviews}
}

@book{lesieur2005large,
  title={Large-eddy simulations of turbulence},
  author={Lesieur, Marcel and M{\'e}tais, Olivier and Comte, Pierre},
  year={2005},
  publisher={Cambridge university press}
}

@article{lesieur1996new,
  title={New trends in large-eddy simulations of turbulence},
  author={Lesieur, Marcel and Metais, Olivier},
  journal={Annual review of fluid mechanics},
  volume={28},
  number={1},
  pages={45--82},
  year={1996},
  publisher={Annual Reviews 4139 El Camino Way, PO Box 10139, Palo Alto, CA 94303-0139, USA}
}

@article{speziale1991analytical,
  title={Analytical methods for the development of Reynolds-stress closures in turbulence},
  author={Speziale, Charles G},
  journal={Annual review of fluid mechanics},
  volume={23},
  number={1},
  pages={107--157},
  year={1991},
  publisher={Annual Reviews 4139 El Camino Way, PO Box 10139, Palo Alto, CA 94303-0139, USA}
}

@article{duraisamy2021perspectives,
  title={Perspectives on machine learning-augmented Reynolds-averaged and large eddy simulation models of turbulence},
  author={Duraisamy, Karthik},
  journal={Physical Review Fluids},
  volume={6},
  number={5},
  pages={050504},
  year={2021},
  publisher={APS}
}

@article{launder1975progress,
  title={Progress in the development of a Reynolds-stress turbulence closure},
  author={Launder, Brian Edward and Reece, G Jr and Rodi, W},
  journal={Journal of fluid mechanics},
  volume={68},
  number={3},
  pages={537--566},
  year={1975},
  publisher={Cambridge University Press}
}

@article{speziale1991modelling,
  title={Modelling the pressure--strain correlation of turbulence: an invariant dynamical systems approach},
  author={Speziale, Charles G and Sarkar, Sutanu and Gatski, Thomas B},
  journal={Journal of fluid mechanics},
  volume={227},
  pages={245--272},
  year={1991},
  publisher={Cambridge University Press}
}

@article{moin1997tackling,
  title={Tackling turbulence with supercomputers},
  author={Moin, Parviz and Kim, John},
  journal={Scientific American},
  volume={276},
  number={1},
  pages={62--68},
  year={1997},
  publisher={JSTOR}
}

@article{catalano2010turbulence,
  title={Turbulence modeling for low-Reynolds-number flows},
  author={Catalano, P and Tognaccini, Renato},
  journal={AIAA journal},
  volume={48},
  number={8},
  pages={1673--1685},
  year={2010}
}

@article{rapp2011flow,
  title={Flow over periodic hills: an experimental study},
  author={Rapp, Ch and Manhart, M},
  journal={Experiments in fluids},
  volume={51},
  number={1},
  pages={247--269},
  year={2011},
  publisher={Springer}
}

@article{moin1998direct,
  title={Direct numerical simulation: a tool in turbulence research},
  author={Moin, Parviz and Mahesh, Krishnan},
  journal={Annual review of fluid mechanics},
  volume={30},
  number={1},
  pages={539--578},
  year={1998},
  publisher={Annual Reviews 4139 El Camino Way, PO Box 10139, Palo Alto, CA 94303-0139, USA}
}

@article{moser1999direct,
  title={Direct numerical simulation of turbulent channel flow up to Re$\tau$= 590},
  author={Moser, Robert D and Kim, John and Mansour, Nagi N},
  journal={Physics of fluids},
  volume={11},
  number={4},
  pages={943--945},
  year={1999},
  publisher={AIP Publishing}
}

@article{lee2015direct,
  title={Direct numerical simulation of turbulent channel flow up to},
  author={Lee, Myoungkyu and Moser, Robert D},
  journal={Journal of fluid mechanics},
  volume={774},
  pages={395--415},
  year={2015},
  publisher={Cambridge University Press}
}

@article{wood1990modeling,
  title={Modeling imprecision and uncertainty in preliminary engineering design},
  author={Wood, Kristin L and Antonsson, Erik K},
  journal={Mechanism and Machine Theory},
  volume={25},
  number={3},
  pages={305--324},
  year={1990},
  publisher={Elsevier}
}

@article{gurnani2005robust,
  title={Robust multiattribute decision making under risk and uncertainty in engineering design},
  author={Gurnani, Ashwin P and Lewis, Kemper},
  journal={Engineering Optimization},
  volume={37},
  number={8},
  pages={813--830},
  year={2005},
  publisher={Taylor \& Francis}
}

@article{du2000methodology,
  title={Methodology for managing the effect of uncertainty in simulation-based design},
  author={Du, Xiaoping and Chen, Wei},
  journal={AIAA journal},
  volume={38},
  number={8},
  pages={1471--1478},
  year={2000}
}

\end{document}